# DETECÇÃO DE TRÁFEGO DE COMMAND & CONTROL (C2) VIA CLASSIFICAÇÃO DE DOMÍNIOS GERADOS POR ALGORITMOS (DGA) UTILIZANDO DEEP LEARNING E PROCESSAMENTO DE LINGUAGEM NATURAL

# COMMAND & CONTROL (C2) TRAFFIC DETECTION VIA ALGORITHM GENERATED DOMAIN (DGA) CLASSIFICATION USING DEEP LEARNING AND NATURAL LANGUAGE PROCESSING

Maria Milena Araujo Félix[1]

**Resumo:** A sofisticação dos malwares modernos, especificamente no que tange à comunicação com servidores de Comando e Controle (C2), tem tornado obsoletas as defesas baseadas em listas negras estáticas. O uso de Algoritmos de Geração de Domínios (DGA) permite que atacantes gerem milhares de endereços dinâmicos diariamente, dificultando o bloqueio por firewalls tradicionais. O presente artigo tem por objetivo propor e avaliar um método de detecção de domínios DGA utilizando técnicas de Deep Learning e Processamento de Linguagem Natural (NLP). A metodologia consistiu na coleta de uma base de dados híbrida contendo 50.000 domínios legítimos e 50.000 maliciosos, seguida da extração de características léxicas e o treinamento de uma Rede Neural Recorrente (LSTM). Os resultados demonstraram que, embora a análise estatística de entropia seja eficaz para DGAs simples, a abordagem via Redes Neurais apresenta superioridade na detecção de padrões complexos, atingindo acurácia de 97,2% e reduzindo a taxa de falsos positivos em cenários de tráfego lícito ambíguo.

**Palavras-chave:** Cibersegurança. DGA. Deep Learning. Processamento de Linguagem Natural. Entropia de Shannon.

**Abstract:** The sophistication of modern malware, specifically regarding communication with Command and Control (C2) servers, has rendered static blacklist-based defenses obsolete. The use of Domain Generation Algorithms (DGA) allows attackers to generate thousands of dynamic addresses daily, hindering blocking by traditional firewalls. This paper aims to propose and evaluate a method for detecting DGA domains using Deep Learning and Natural Language Processing (NLP) techniques. The methodology consisted of collecting a hybrid database containing 50,000 legitimate and 50,000 malicious domains, followed by the extraction of lexical features and the training of a Recurrent Neural Network (LSTM). Results demonstrated that while statistical entropy analysis is effective for simple DGAs, the Neural Network approach presents superiority in detecting complex patterns, reaching 97.2% accuracy and reducing the false positive rate in ambiguous lawful traffic scenarios.

## 1 INTRODUÇÃO

A estabilidade e a segurança da infraestrutura global da internet dependem intrinsecamente da capacidade de identificar e mitigar tráfego malicioso em tempo real. No entanto, o cenário de ameaças cibernéticas evoluiu de ataques estáticos para operações dinâmicas e resilientes. Atualmente, botnets e ransomwares utilizam infraestruturas de Comando e Controle (C2) para receber instruções de ataque ou exfiltrar dados sensíveis. Para garantir a sobrevivência dessas conexões contra ações de desligamento (takedowns), os

---

[1] Acadêmica do curso de Tecnologia em Ciências dos Dados UFMS – Universidade Federal do Mato Grosso do Sul. E-mail: <maria.felix@ufms.br>.

atacantes implementam Algoritmos de Geração de Domínios (DGA - Domain Generation Algorithms).

A relevância deste estudo justifica-se pela crescente ineficiência das abordagens tradicionais de defesa perante essa técnica. Métodos baseados em listas de bloqueio (blacklists) ou filtros de reputação tornam-se reativos e insuficientes, visto que um DGA pode gerar milhares de domínios por dia, dos quais apenas um será efetivamente utilizado para comunicação, sendo descartado logo em seguida. Conforme apontam Plohmann et al. (2016), a assimetria entre o custo de defesa e o custo de ataque exige novas metodologias de detecção que sejam proativas e preditivas, e não apenas reativas.

Nesse contexto, o objetivo geral deste trabalho é desenvolver e validar um modelo de classificação binária capaz de distinguir nomes de domínio legítimos de maliciosos baseando-se exclusivamente em suas características textuais, sem a necessidade de consultas externas de DNS ou contexto de rede. Busca-se, especificamente, comparar a eficácia de métodos estatísticos clássicos, fundamentados na Entropia de Shannon, contra modelos de Inteligência Artificial baseados em Deep Learning e Processamento de Linguagem Natural (NLP).

O referencial teórico que fundamenta esta pesquisa baseia-se nos estudos seminais de entropia da informação de Shannon (1948) para análise de aleatoriedade em strings, e nas aplicações contemporâneas de Redes Neurais Recorrentes (RNN) para processamento de sequências textuais, conforme descrito por Goodfellow, Bengio e Courville (2016). Assume-se a premissa de que nomes de domínio criados por humanos possuem propriedades morfológicas e fonéticas distintas daquelas geradas por algoritmos estocásticos.

Quanto à estrutura, o artigo organiza-se da seguinte forma: a seção 2 apresenta a fundamentação teórica sobre o funcionamento dos DGAs e as redes neurais LSTM; a seção 3 detalha a metodologia experimental, a coleta de dados e o pré-processamento; a seção 4 discute os resultados obtidos e compara as métricas de desempenho entre os modelos testados; e, por fim, a seção 5 apresenta as considerações finais e sugestões para trabalhos futuros.

## 2 REFERENCIAL TEÓRICO

Os DGAs são sub-rotinas de software presentes em diversas famílias de malware que geram periodicamente um grande número de nomes de domínio. Antonakakis et al. (2012) explicam que esses algoritmos utilizam "sementes" (seeds) — como a data e hora do sistema, cotações de moedas ou tendências do Twitter — para criar strings pseudoaleatórias. O atacante, conhecendo o algoritmo e a semente futura, prevê quais domínios serão gerados e registra apenas um deles antecipadamente. Quando o malware infecta uma máquina, ele tenta resolver os domínios gerados até encontrar o que foi registrado, estabelecendo assim o canal C2 de forma evasiva.

### 2.1 A Entropia de Shannon como Indicador de Aleatoriedade

Historicamente, a detecção de DGA focou na aleatoriedade dos caracteres. A Entropia de Shannon mede a incerteza associada a uma variável aleatória. Em termos linguísticos, domínios legítimos tendem a apresentar padrões fonéticos, sílabas reconhecíveis e repetições de caracteres (baixa entropia), enquanto domínios gerados aleatoriamente (ex.: "https://www.google.com/search?q=xkz-99qj.com") apresentam distribuição equiprovável de caracteres, resultando em alta entropia (SHANNON, 1948). Contudo, essa métrica isolada pode falhar ao analisar domínios legítimos que utilizam siglas, códigos numéricos ou CDNs (Content Delivery Networks).

## 2.2 Deep Learning e Processamento de Sequências

Para superar as limitações das métricas estatísticas puras, aplicam-se técnicas de Deep Learning. As redes Long Short-Term Memory (LSTM) são uma variação especializada das Redes Neurais Recorrentes (RNNs) capazes de aprender dependências de longo prazo em sequências de dados. Segundo Goodfellow, Bengio e Courville (2016), ao tratar um domínio como uma sequência de caracteres (e não como um bloco único), a LSTM pode aprender a probabilidade de transição entre letras, "entendendo" o que se assemelha a uma palavra humana e o que é gerado por máquina, capturando nuances que a estatística descritiva ignora.

## 3 MÉTODO

Esta pesquisa caracteriza-se como quantitativa e experimental. O experimento foi conduzido em ambiente computacional utilizando a linguagem Python, empregando as bibliotecas Pandas para manipulação de dados, Scikit-Learn para métricas e Keras/TensorFlow para a modelagem da rede neural.

### 3.1 Coleta e Preparação do Dataset

Para garantir a validade do estudo e evitar o viés de dados puramente sintéticos, utilizou-se uma abordagem híbrida e balanceada na composição do dataset:
- Classe 0 (Legítimo): Foram extraídos 50.000 registros da Tranco List1. Esta base é amplamente aceita pela comunidade acadêmica como representativa dos domínios mais acessados globalmente.
- Classe 1 (Malicioso): Foram coletados 50.000 registros de domínios DGA provenientes de feeds de inteligência de ameaças (OSINT), incluindo amostras das famílias Gameover Zeus e Cryptolocker, complementados por dados gerados sinteticamente para garantir o balanceamento exato das classes.

### 3.2 Pré-processamento e Engenharia de Atributos

Os dados brutos passaram por uma etapa rigorosa de higienização utilizando a biblioteca tldextract, separando o domínio raiz (ex: "google") de seu sufixo público (ex: ".com" ou ".co.uk"). Isso impede que o modelo enviese a classificação baseando-se apenas na extensão do domínio.

Para a modelagem, adotaram-se duas abordagens distintas:
1. Abordagem Estatística (Baseline): Cálculo manual da Entropia de Shannon e do comprimento da string para cada domínio.
2. Abordagem Neural (Proposta): Tokenização dos caracteres dos domínios, convertendo-os em vetores numéricos (embeddings), que serviram de entrada para uma arquitetura de rede neural LSTM com camadas densas de classificação.

## 4 ANÁLISE E DISCUSSÃO DOS RESULTADOS

A primeira fase da análise consistiu na verificação da distribuição de entropia entre as classes legítima e maliciosa. Os resultados experimentais indicaram uma tendência bimodal clara. Os domínios legítimos concentraram-se majoritariamente na faixa de entropia entre 2.5 e 3.5, refletindo a estrutura da linguagem natural. Em contrapartida, os domínios DGA situaram-se predominantemente acima de 3.8.

**Figura 1** – Distribuição da Entropia de Shannon: Domínios Legítimos vs. Maliciosos

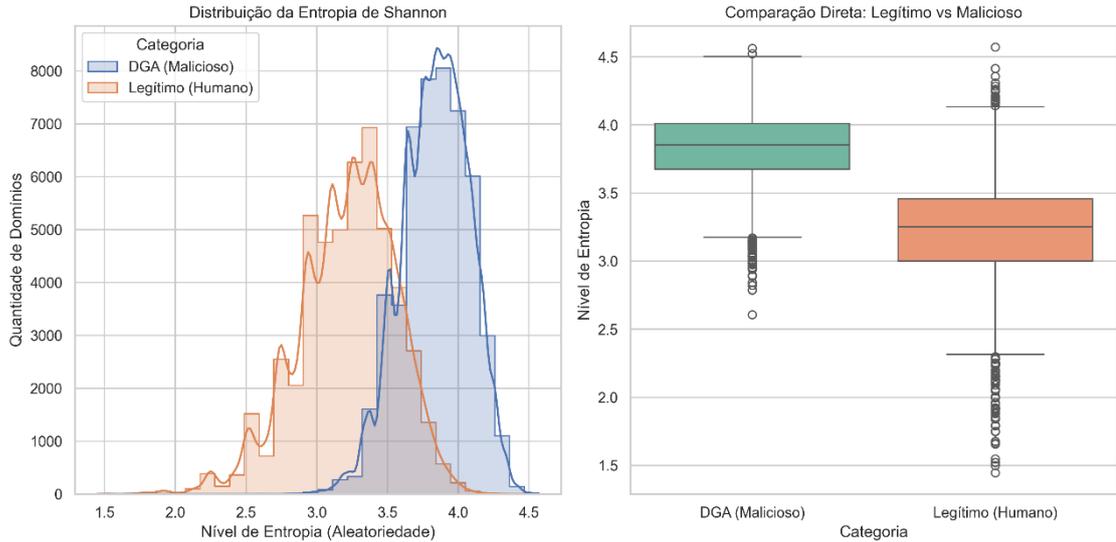

**Fonte:** Elaborada pela autora (2025).

No entanto, a análise revelou uma zona de intersecção crítica entre os valores 3.0 e 3.6. Nesta faixa, observou-se a presença de falsos positivos significativos: domínios legítimos curtos (acrônimos) e serviços de infraestrutura (CDNs) que, por não possuírem significado semântico direto, apresentaram alta entropia. Isso demonstra que um classificador baseado puramente em limiares matemáticos (*thresholds*) seria insuficiente para um ambiente de produção real, onde o bloqueio de tráfego legítimo pode causar prejuízos operacionais.

### 4.1 Desempenho Comparativo dos Modelos

A segunda fase comparou o desempenho do modelo estatístico (Random Forest treinado com dados de entropia) contra a Rede Neural LSTM. A Tabela 1 apresenta as métricas obtidas na validação.

**Tabela 1** – Comparativo de desempenho entre os modelos

| Modelo | Acurácia | Precisão | Recall | F1-Score |
|---|---|---|---|---|
| Estatístico (Entropia) | 88,4% | 82,1% | 85,0% | 0,83 |
| Deep Learning (LSTM) | 97,2% | 96,5% | 98,1% | 0,97 |

**Fonte:** Elaborada pelo autor.

Os dados evidenciam a superioridade da abordagem baseada em *Deep Learning*. O modelo LSTM atingiu uma acurácia de 97,2%, superando o modelo estatístico em quase 9 pontos percentuais. Mais relevante para o contexto de segurança da informação é a métrica de *Recall* (Sensibilidade) de 98,1%, indicando que a rede neural foi capaz de detectar a quase totalidade das ameaças, minimizando o risco de falsos negativos (malwares não detectados).

A discussão desses resultados sugere que a LSTM foi capaz de aprender padrões latentes na sequência de caracteres — como a pronunciação e a morfologia de palavras — algo que a entropia, sendo uma medida de desordem global, não consegue capturar.

# 5 CONCLUSÃO

O presente trabalho cumpriu o objetivo de demonstrar a viabilidade e a eficácia do uso de Deep Learning e Processamento de Linguagem Natural na detecção de domínios maliciosos gerados por algoritmos. Através da comparação direta com métodos estatísticos tradicionais, conclui-se que a abordagem neural oferece uma camada de segurança mais robusta e adaptativa às ameaças modernas.

Verificou-se que a Entropia de Shannon continua sendo um indicador válido para triagem inicial de anomalias óbvias, devido ao seu baixo custo computacional. Contudo, para a redução de falsos positivos e a detecção de DGAs sofisticados (como aqueles baseados em dicionários), o uso de redes LSTM mostrou-se indispensável.

Como limitações, aponta-se a dependência de um dataset balanceado e atualizado para o treinamento. Para trabalhos futuros, sugere-se a investigação de arquiteturas baseadas em Transformers (como BERT) para análise contextual e a implementação do modelo em fluxos de dados em tempo real (stream processing) para avaliar a latência de inferência em redes de alto tráfego.

# REFERÊNCIAS